%% file: main.tex
\documentclass[letterpaper, 10 pt, conference]{ieeeconf}  

\IEEEoverridecommandlockouts                              

\usepackage{graphics} 
\usepackage{epsfig} 
\usepackage{mathptmx} 
\usepackage{times} 
\usepackage{amsmath} 
\usepackage{amssymb}  
\usepackage{subcaption}
\usepackage{tabularx}
\usepackage{color}
\usepackage{xcolor}
\usepackage{hyperref}

\title{\LARGE \bf
Efficient Map Prediction via Low-Rank Matrix Completion 
}

\author{Zheng Chen, Shi Bai, Lantao Liu
\thanks{\newline Zheng Chen and Lantao Liu are with the Luddy School of Informatics, Computing, and Engineering  at Indiana University, Bloomington, IN 47408, USA. E-mail:
        {\tt\small \{zc11, lantao\}@iu.edu}
        \newline Shi Bai is with Wing, Alphabet Inc. E-mail: {\tt\small baishi@wing.com}
        }%
}

\begin{document}
\maketitle
\thispagestyle{empty}
\pagestyle{empty}

\input{0_abstract}
\input{1_introduction}
\input{2_related_work}
\input{3_methodology}

\input{4_experiments}

\input{5_conclusions}

\bibliographystyle{unsrt}
\bibliography{ref}

\end{document}

%% file: 0_abstract.tex
\begin{abstract}
In many autonomous mapping tasks, the maps cannot be accurately constructed due to various reasons such as sparse, noisy, and partial sensor measurements.  
We propose a novel map prediction method built upon recent success of \textit{Low-Rank Matrix Completion}. The proposed map prediction is able to achieve both map interpolation and extrapolation on raw poor-quality maps with missing or noisy observations. 
We validate with extensive simulated experiments that the approach can achieve real-time computation for large maps, and the performance is superior to state-of-the-art map prediction approach --- Bayesian Hilbert Mapping in terms of mapping accuracy and computation time.
Then we demonstrate that with the proposed real-time map prediction framework, the coverage convergence rate~(per action step) for a set of representative coverage planning methods commonly used for environmental modeling and monitoring tasks can be significantly improved.

\end{abstract}

%% file: 1_introduction.tex
\section{INTRODUCTION}
Mapping is a critical functionality for an autonomous robot that needs to perform tasks in an initially unknown environment where an underlying map is constructed with onboard sensors. 
However, this can be challenging if the sensor measurements are inaccurate or if the environment can only be partially measured (sensed), which are commonly seen in autonomous mapping scenarios. 
For instance, the point clouds obtained from low-end proximity sensors (e.g., LiDAR, IR, Sonar) might be extremely sparse especially in spacious environments. The point clouds might also contain many outliers (noisy points). In addition, oftentimes a mapping robot cannot visit every spot of the environment and the traversed areas may only partially cover the space with only some small sub-regions being measured. 
Given the sparse, noisy, and partial observations, it is desirable that a complete map can be efficiently predicted to well match the ground-truth environment, which is the objective of this paper. 

More formally, the map prediction aims to predict map structure based on already known regions and could be categorized into two classes: interpolation and extrapolation. Map interpolation means given some observations (measurement samples), we want to predict the values \textit{in between} the sampled places;
Map extrapolation means given some observation samples (or map structures), we want to predict the map values \textit{beyond} the sampled places. 

\begin{figure} \vspace{5pt}
  \centering
  	{\label{fig:real}\includegraphics[width=\linewidth]{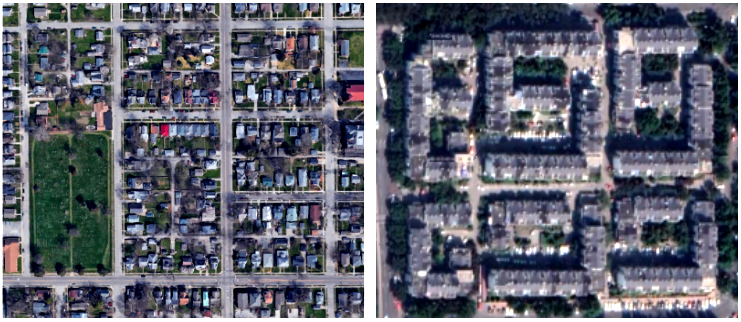}}
  \caption{\small Urban and residential environments reveal strong structured patterns, such as \textit{Left}: road (or street) network and \textit{Right}: buildings layout. These environments containing linear dependent structures can be modeled as low-rank-matrix maps.} 
\label{fig:real}  
\end{figure}

In this paper, we propose to use Low-Rank Matrix Completion~(LRMC) to leverage the ill-conditioned map data to predict the whole map. LRMC exploits some special structures, such as low rank~(or linear dependency) and incoherence hidden in underlying map to perform the prediction. Two examples that possess special patterns in urban/residential environments are shown in Fig.~\ref{fig:real}.
We first illustrate how a structured environmental map satisfies important assumptions required by LRMC model. Then we qualitatively demonstrate with an urban road network map that our proposed LRMC method outperforms state-of-the-art map prediction~(or regression) method -- Bayesian Hilbert Mapping~(BHM)~\cite{senanayake2017bayesian} which is a Bayesian extension of Hilbert Mapping~(HM)~\cite{ramos2016hilbert}, where the map prediction is treated as a kernelized logistic regression problem. BHM eliminates crucial regularization parameter tunning in HM, and outputs continuous maps where the occupancy value of any arbitrary point could be queried.

We conduct exhaustive simulations on a set of maze-like complex maps and quantitatively analyze that our proposed LRMC method is superior to BHM on the maps with varying linear patterns and different rank values. 
Finally we combine the proposed real-time map prediction with representative coverage planning methods commonly used for environmental mapping and monitoring, and show that the mapping coverage convergence could be significantly improved with high mapping accuracy. 

We summarize our contributions as follows:
\begin{itemize}
    \item 
    This is the first time to examine and report that many complex urban/residential environments possess low-rank and incoherent structure, and to apply Low-Rank Matrix Completion for map prediction based on sparse, noisy, and partially observed maps.
    \item
    Our proposed Low-Rank Matrix Completion based map prediction outperforms state-of-the-art map prediction method---Bayesian Hilbert Mapping in terms of mapping accuracy and computation time and is able to perform prediction in real-time.
    \item
    We perform extensive simulations and demonstrate the remarkable effectiveness of our proposed method  which allows representative coverage planning methods to achieve faster mapping coverage convergence rates.  
\end{itemize}

%% file: 2_related_work.tex
\section{RELATED WORK}

Map prediction has been studied in occupancy mapping problems.  
Prevalent methods include Occupancy Girds Mapping~(OGM)~\cite{elfes1989using}, Gaussian Process Occupancy Mapping~(GPOM)~\cite{o2012gaussian, wang2016fast}, Hilbert Mapping~(HM)~\cite{ramos2016hilbert} and Bayesian Hilbert Mapping~(BHM)~\cite{senanayake2017bayesian}. Specifically, OGM belongs to discrete mapping and all grids are independent with each other. To eliminate the requirement of pre-defining discrete grids and the independence assumption in OGM, GPOM is proposed and used in many applications~\cite{jadidi2014exploration, jadidi2015mutual}. GPOM naturally captures the neighbouring information thanks to kernelization, but being a non-parametric model, it has a computation complexity of $O(n^3)$, where $n$ is the number of sampled data points. Combining various advantages of GPOM, a faster and simpler parametric model---HM and its Bayesian version, BHM, is proposed later to eliminate the cubical time complexity in GPOM.

Map extrapolation 
attracts increasing attention in recent years and almost all the current methods for map extrapolation are based on deep neural networks~(DNNs)~\cite{caley2019deep, pronobis2017learning, katyal2018occupancy, shrestha2019learned, saroya2020online, katyal2019uncertainty}. However, DNNs based methods require huge amount of data for training and hence are still limited in general applications. In this paper, we propose to use Low-Rank Matrix Completion to perform the map prediction. 
Our proposed LRMC based method falls in between the {interpolation} and {extrapolation} methods. Specifically, the LRMC method can not only well interpolate map values, but at the same time can  extrapolate the map structure beyond the explored areas. 

In the past two decades, the Low-Rank Matrix Completion~(LRMC) problem has been well studied. The first theoretically guaranteed exact LRMC algorithm is proposed in~\cite{candes2009exact}, where any $n \times n$ incoherent matrices of rank $r$ are proven to be exactly recovered from $C n^{1.2}r\log n$ uniformly randomly sampled entries with high probability through solving a convex problem of nuclear norm minimization~(NNM). Subsequent works~\cite{candes2010power, chen2015incoherence, gross2011recovering, recht2011simpler} refine the provable completion results following the NNM based method. However, since all of the algorithms mentioned above are based on second order methods~\cite{liu2010interior}, they can become extremely expensive if the matrix dimension is large~\cite{cai2010singular}. Some first order based methods~\cite{cai2010singular, ji2009accelerated, mazumder2010spectral} are developed later. They solve the nuclear norm minimization problem in an iterative manner and rely on Singular Value Decomposition~(SVD) of matrices and are suited to large-scale matrix completion problems. In recent years, many other techniques are developed, such as coherent matrix completion~\cite{liu2017new, bhojanapalli2014universal, chen2014coherent}, non-linear matrix completion~\cite{eriksson2012high} and adaptive sampling~\cite{chen2015completing, eftekhari2018mc2}. {\color{black}A recent survey on LRMC could be found in \cite{nguyen2019low}.}
In this paper, we mainly employ the first order SVD based iterative method, such as~\cite{mazumder2010spectral} to solve the LRMC problem. {\color{black} The SVD based methods are able to achieve low prediction error and maintain a low time complexity if certain special problem structure is appropriately leveraged~\cite{mazumder2010spectral}.}

%% file: 3_methodology.tex
\section{METHODOLOGY}
\label{sect:method}

\subsection{Problem Description}
Map prediction exhibits the same form as the matrix completion \cite{recht2009simpler} while a 2D map can be treated as a matrix. We propose to use LRMC as the map prediction method. LRMC is defined as completing a partially observed matrix $\tilde{\mathbf{M}}\in\mathbb{R}^{n_1\times n_2}$ whose part of entries are missing and the corresponding ground-truth matrix $\mathbf{M}$ has a low-rank and incoherent structures. Since the ground-truth rank is usually unknown, a typical way to complete $\tilde{\mathbf{M}}$ is to find a minimal rank matrix $\mathbf{X}^*$ that is consistent with $\tilde{\mathbf{M}}$ on the observed entries. A formal mathematical definition is:
\begin{equation}
    \label{eq:mc_1}
    \begin{aligned}
        \mathbf{X}^* = ~&\underset{\mathbf{X}}{\text{argmin}} 
               &&rank(\mathbf{X})\\
               &\text{subject to}
               && X_{ij} = M_{ij},~~\left ( i, j \right )\in \Omega,
    \end{aligned}
\end{equation}
where $\Omega \subset \left \{ 1, \cdots, n_1 \right \}\times \left \{ 1, \dots, n_2 \right \} $ is an index set of locations corresponding to the observed entries~($\left ( i, j \right )\in \Omega$ if $M_{ij}$ is observed). 

\subsection{Low-Rank and Incoherent Structures}
Two important assumptions for the underlying matrix $\mathbf{M}$ must be satisfied to guarantee the matrix completion results:
\begin{itemize}
    \item 
    $\mathbf{M}$ should be a low-rank matrix. A low-rank matrix implies its rank value is much smaller than its dimension and this structure enables the possibility to leverage linear dependencies among columns and/or rows of a matrix to predict missing entries~\cite{ongie2018tensor}. 
    \item
    $\mathbf{M}$ should be an incoherent matrix, which means the coherence of $\mathbf{M}$ is low. Suppose that a \textit{compact} SVD of a rank-r matrix $\mathbf{M}\in \mathbb{R}^{n_1\times n_2}$ is $\mathbf{M} = \mathbf{U}\Sigma\mathbf{V}^T$, where $\mathbf{U} \in \mathbb{R}^{n_1\times r}$ and $\mathbf{V}\in \mathbb{R}^{n_2\times r}$ are stacked left and right singular vectors of $\mathbf{M}$, while $\Sigma \in \mathbb{R}^{r\times r}$ is the diagonal matrix formed by a number of $r$ singular values of $\mathbf{M}$. The statistical leverage scores of $\mathbf{U}$ and $\mathbf{V}$ are defined by $\gamma_{U} = \underset{i\in \left \{ 1, \cdots, n_1 \right \}}{\text{max}}\left \| \mathbf{U}_{i,~:} \right \|^2_2$ and $\gamma_{V} = \underset{j\in \left \{ 1, \cdots, n_2 \right \}}{\text{max}}\left \| \mathbf{V}_{j,~ :} \right \|^2_2$, respectively, where $\mathbf{U}_{i,~:}$ and $\mathbf{V}_{j,~:}$ represent the $i^{th}$ and $j^{th}$ row of $\mathbf{U}$ and $\mathbf{V}$. Then the coherence of $\mathbf{M}$ is defined as:
    \begin{equation}
        \label{eq:coherence}
        \gamma_{M} = \max \left \{ \gamma_{U}, \gamma_{V} \right \}.
    \end{equation}
    The value of $\gamma_M$ is within $\left [ 0, 1 \right ]$ and an incoherent structure means the value of $\gamma_M$ is near 0. A small coherence value implies that the mass of values in the matrix will not concentrate on small portions, but spread out through the whole matrix.
\end{itemize}

Under the two conditions, the matrix $\tilde{\mathbf{M}}$ has been shown to be provably recoverable~\cite{candes2009exact}. Fig.~\ref{fig:mazes_1} and Table~\ref{tb:maze_prop} illustrate the trends of ranks and coherences of different structured environments which are similar to complex versions of either urban road networks or building layouts.
We can observe that the typical structured environments have ranks below 100, which are significantly smaller compared to the original matrix dimensions which could be multiple magnitudes higher (e.g., $10^4$ or above). 
The trend of ranks also reflects the complexity of the environment, and the higher the rank is, the more complex the environment is.
Along with the rank values, we also list the coherence values computed by using Eq.~(\ref{eq:coherence}).
Although the coherence increases proportionally to the rank, they all stay at a low level indicating high incoherence. 

\begin{figure} \vspace{-3pt}
  \centering
  	{\label{fig:mazes_1}\includegraphics[width=0.99\linewidth]{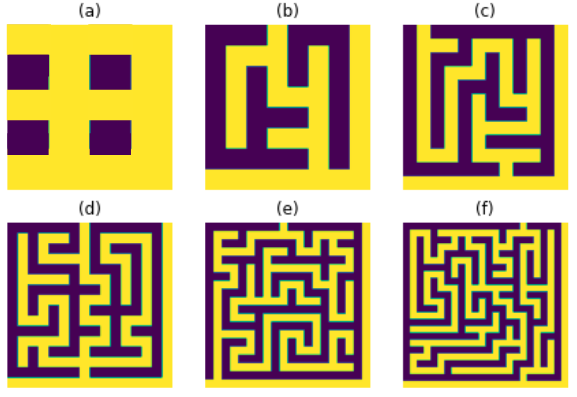}}
  \caption{\small Maze environments with different ranks.
  } \vspace{-10pt}
\label{fig:mazes_1}  
\end{figure}

\begin{table}[h]

\caption{Matrix properties of maze environments} 
\centering 
\begin{tabularx}{\linewidth}{ccccccc}
\hline\hline 
   & (a) & (b) & (c) & (d) & (e) & (f)\\
\hline 
Rank & 2 & 7 & 11 & 14 & 19 & 22\\ 
Coherence & 0.0200 & 0.0400 & 0.0625 & 0.0833 & 0.1000 & 0.1250\\ 
\hline 
\end{tabularx}
\label{tb:maze_prop}
\end{table}

\subsection{Solving LRMC in Map Prediction}
\label{sect:static}

To solve the LRMC problem, 
the rank minimization problem in Eq.~(\ref{eq:mc_1}) is non-convex and NP-hard unfortunately, and hence of little practical use. An alternative way for Eq.~(\ref{eq:mc_1}) is to replace $rank(\mathbf{X})$ with $\left \| \mathbf{X} \right \|_*$, where $\left \| \mathbf{X} \right \|_*$ is the nuclear norm of $\mathbf{X}$, that is, the sum of sigular values of $\mathbf{X}$: $\left \| \mathbf{X} \right \|_* = \sum_{k=1}^{\min \left \{ n_1, n_2 \right \}}\sigma_k(\mathbf{X})$, where $\sigma_k(\mathbf{X})$ is the $k^{th}$ largest singular values of $\mathbf{X}$. The new formulation is:
\begin{equation}
    \label{eq:mc_2}
    \begin{aligned}
        \mathbf{X}^* = ~&\underset{\mathbf{X}}{\text{argmin}} 
               &&\left \| \mathbf{X} \right \|_*\\
               &\text{subject to}
               && X_{ij} = M_{ij},~~\left ( i, j \right )\in \Omega.
    \end{aligned}
\end{equation}
The nuclear norm $\left \| \mathbf{X} \right \|_*$ is an effective convex relaxation to the rank objective and the optimization in Eq.~(\ref{eq:mc_2}) could be solved by semi-definite programming~\cite{laurent2014positive}.

Nevertheless, the optimizations in Eq.~(\ref{eq:mc_1}) and Eq.~(\ref{eq:mc_2}) aim to exactly recover the partially observed matrix, and this way was claimed as too rigid and may result in over-fitting~\cite{mazumder2010spectral}. A more robust way is to add a regularization parameter in Eq.~(\ref{eq:mc_2}):
\begin{equation}
    \label{eq:mc_3}
    \begin{aligned}
        \mathbf{X}^* = ~&\underset{\mathbf{X}}{\text{argmin}} 
               &&\left \| \mathbf{X} \right \|_*\\
               &\text{subject to}
               && \sum_{(i, j)\in \Omega}(X_{ij} - M_{ij})^2\leq \delta,
    \end{aligned}
\end{equation}
where $\delta \geq 0$ is a parameter regulating the training error tolerance. Equivalently Eq.~(\ref{eq:mc_3}) could be reformulated in a Lagrange form:
\begin{equation}
    \label{eq:mc_4}
    \begin{aligned}
        \mathbf{X}^* = \underset{\mathbf{X}}{\text{argmin}} \sum_{(i, j)\in \Omega} (X_{ij} - M_{ij})^2 + \lambda \left \|\mathbf{X} \right \|_{*},
    \end{aligned}
\end{equation}
where $\lambda$ is a regularization parameter controlling the nuclear norm. A SVD based iterative method proposed in~\cite{mazumder2010spectral} can be used to solve the Eq.~(\ref{eq:mc_4}).

\subsection{Mapping with Non-myopic Planner}
\label{sect:plan}

We are intrigued if the proposed map prediction can facilitate and improve existing environmental mapping and coverage methods. Popular planning paradigms include lawnmower mapping~\cite{li2019coverage, song2018varepsilon}, myopic (short-horizon) planning~\cite{khan2014greedy}, as well as non-myopic (long-horizon) planning~\cite{ kantaros2019asymptotically}. Lawnmower planning pursues complete spatial coverage, and myopic methods are oftentimes greedy schemes. The non-myopic approaches have been favored by many researchers for environmental mapping and monitoring~\cite{ma2016information, chen2019pareto}. Since non-myopic planners optimize paths over some horizon, we adopt and adapt the standard  \textit{Traveling Salesman Problem}~(TSP), which will be used in our evaluation in the experimental section. 

A TSP 
can be defined as that, given the positions of a collection of way points, what is the shortest possible path of reaching each way point only once and eventually returning to the starting point. 
To efficiently solve the TSP, we extend our recently developed adaptive \textit{k}-opt method~\cite{ma2016adaptive} which exploits and modifies the adaptive $k$-swap mechanism originally designed for multi-agent
task allocation. The core idea is to approximate the NP-hard problem with an opportunistic and localized tour improvement heuristic by taking account of tour
constraints adaptively and flexibly, so that real-time solution can be obtained with small sacrifice on solution accuracy. 

To determine the number of the way points for navigation guidance in TSP, previous work~\cite{candes2009exact} has theoretically proven that a partially observed low-rank matrix could be perfectly recovered if the number of sampled entries $m$ obeys:
\begin{equation}
    \label{eq:sampled_cond}
        m\geq Cn^{1.2}r\log n,
\end{equation}
where $C$ is a positive numerical coefficient, $n = \max\left \{ n_1, n_2 \right \}$, and $r$ is the rank of the underlying matrix $\mathbf{M}$~\cite{candes2009exact}. Based on Eq.~(\ref{eq:sampled_cond}),
the placements of way points
can be defined
as:
\begin{equation}
    \label{eq:sample_num}
    N^{'}_{s} = \left \lfloor \frac{C\cdot n^{1.2}r\log n}{N} \right \rfloor
\end{equation}
where $N^{'}_{s}$ represents the number of way points and $N$ is the number of observed map samples within each way point's local sensing range. 
Note that, in coverage planning scenarios, the robot can take observations in between any pair of successive sampling way points and may introduce additional observations. This implies we may want to reduce the number of sampled way points, 
and we do this by using a scale factor $\epsilon$, so that $N_{s} = \left \lfloor \epsilon N^{'}_{s} \right \rfloor$, 
where $N_{s}$ is the number of sampled way points in TSP planning. 
We name this TSP based planning with different $\epsilon$ values as $TSP\_\epsilon$. 
Once a number of $N_{s}$ points are sampled, a fast and  near-shortest path will be formed using the adaptive \textit{k}-opt TSP solver and thereafter our robot could start the coverage mission by following the path.

%% file: 4_experiments.tex
\section{EXPERIMENTS}

To validate the proposed framework, we perform extensive simulation evaluations with a variety of complex maps. 

\subsection{Experimental Setups and Metrics Design}

To account for the sensor limitations as in robotics applications, we use two measurement patterns, \textit{Noisy Observation}~(NO) and \textit{Partial Observation}~(PO), for constructing the $\Omega$. NO means the observations are imperfect due to random noises (e.g., due to factors of environments/weathers such as snows, or sensor hardware limitations such as sparse and noisy measurements); 
PO means missed or uncovered regions which are typical in environmental mapping and monitoring scenarios. 

To quantitatively compare the two methods,  we use two evaluation metrics, \textit{total prediction accuracy}~(TPA), and \textit{computational time}~(CT). TPA indicates the correct entry matches (including the observed entries) between the entire predicted map and the ground-truth map while CT measures the required time for map prediction computation. 
In addition, we also evaluate the \textit{mapping convergence rate} during the coverage planning.

\subsection{Demonstration with Road Network Mapping}
\begin{figure}
  \centering
  	{\label{fig:road_image}\includegraphics[width=\linewidth]{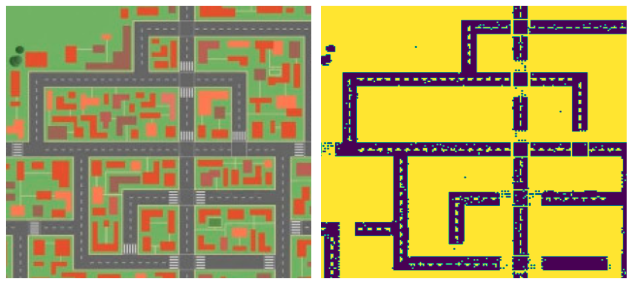}}
  \caption{\small \textit{Left}: road network displays linear dependent structures.  \textit{Right}: The road network is extracted where 
  the dark color represents the road while others are off-road places.
  } 
\label{fig:road_image}  
\end{figure}

\begin{figure} 
  \centering
  	{\label{fig:road_comparison}\includegraphics[width=\linewidth]{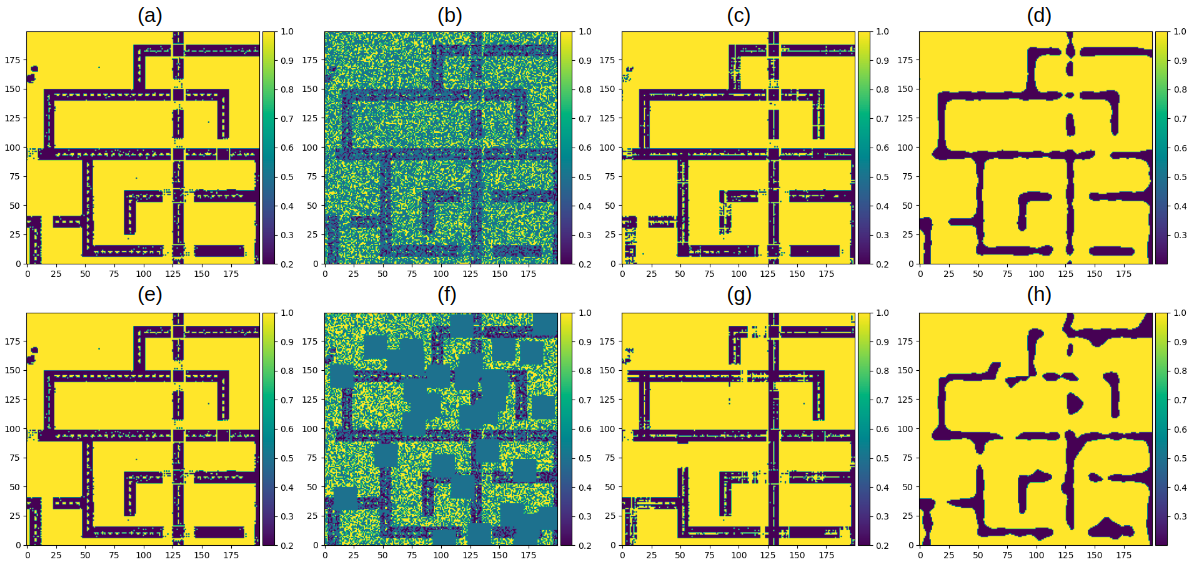}}
  \caption{\small Comparison between LRMC and BHM under NO~(first row) and PO~(second row) patterns, respectively. (a)\&(e)~Ground-truth maps. (b)\&(f)~Partially observed maps for NO and PO patterns. The missing parts are indicated in green patches. (c)\&(g)~Predicted maps from LRMC model. (d)\&(h)~Predicted maps from BHM.
  } \vspace{-10pt}
\label{fig:road_comparison}  
\end{figure}

We first demonstrate that our proposed LRMC method is able to perform map interpolation and extrapolation on a road network map extracted from an  urban environmental map, as shown in Fig.~\ref{fig:road_image}. 
The road surface can be detected by proximity sensors such as IR. We assume the robot is an aerial vehicle so that it can spot regions locally below it. 

The comparison of the two mapping methods on the urban road map under NO and PO patterns can be seen in Fig.~\ref{fig:road_comparison}. With NO pattern, both methods can generally predict the road network pattern~(topology), although BHM is more sensitive to the noises and thus fails to capture the exact road shape. On the other hand, with PO pattern, LRMC significantly outperforms BHM in the presence of multiple missing patches. 
Different from NO pattern, PO requires that the map prediction not only has interpolation capability, but also can perform slight extrapolation to reason the map structure within missing patches. 
LRMC leverages the linear dependency among the rows/columns of the map matrix and is able to predict the structure even multiple missing patches occur. In contrast, although BHM can capture local neighbouring spatial relationship by using kernelization, it is unable to capture the linear dependency brought by the low-rank structure, and thus performs worse than LRMC. 

\subsection{Quantitative Comparison between LRMC and BHM}

We first generate 20 differing complex maze-like layouts/environments by using the Daedalus tool~\cite{daedalus}, and consider the TPA and CT statistical comparisons under NO and PO patterns for different values of $C$. 
The results are shown in Fig.~\ref{fig:us_acc_time_curve} and Fig.~\ref{fig:bus_acc_time_curve}. 
The trend of TPA and CT in different rank-valued maps 
are shown in Fig.~\ref{fig:us_ranks_curve} for NO and Fig.~\ref{fig:bus_ranks_curve} for PO, respectively.

We first analyze the results for NO pattern. In Fig.~\ref{fig:us_acc_time_curve}-\textit{Left}, it can be observed that 
LRMC could achieve a higher accuracy level. 
From Fig.~\ref{fig:us_acc_time_curve}-\textit{Left}, we want to determine a proper scale of the coefficient $C$. We can think the coefficient $C$ is actually reflecting the number of observed map samples according to Eq.~(\ref{eq:sampled_cond}). 
In Fig.~\ref{fig:us_acc_time_curve}-\textit{Left}, we can see that TPA could reach a high level and stay stable when $C\geq 1.5$.
We then compare the CT performance as shown in Fig.~\ref{fig:us_acc_time_curve}-\textit{Right}. The result clearly distinguishes the two methods. The required computation time for LRMC is much lower than that of BHM. Given the size of the entire map~(matrix), the computational complexity of the LRMC algorithm~(SoftImpute~\cite{mazumder2010spectral}) we use is $O((m+n)r^2)$, where $m$ and $n$ are the dimensions and $r$ is the rank of the underlying ground-truth matrix. This means the LRMC model has a constant time complexity regardless of the number of observations. BHM is a logistic regression based method and has linear time complexity in terms of the number of observations. 

\begin{figure} 
  \centering
  	{\label{fig:us_acc_time_curve}\includegraphics[width=0.99\linewidth]{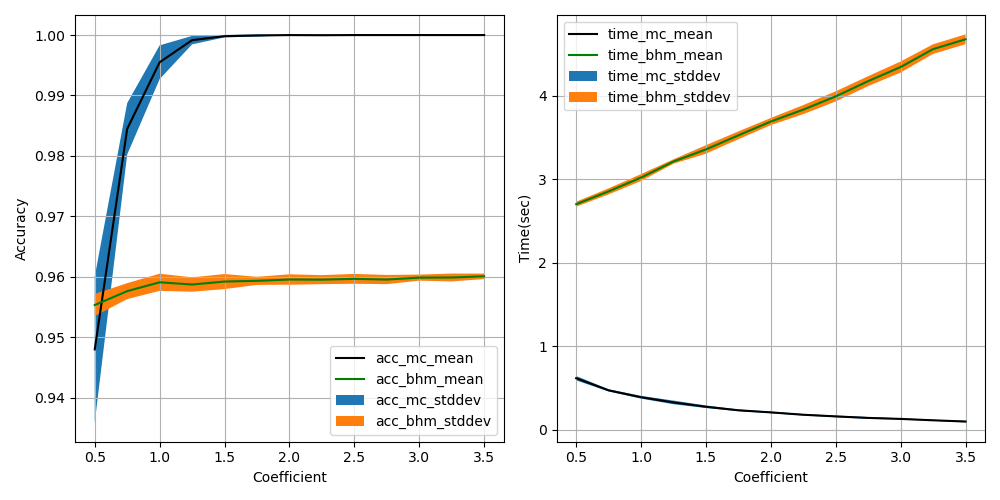}}
  \caption{\small \textit{Left}: Statistical TPA performance and \textit{Right}: CT performance per value of $C$ for LRMC and BHM under NO pattern from 20 different mazes~(with different linear dependencies but the same rank). 
  } \vspace{-10pt}
\label{fig:us_acc_time_curve}  
\end{figure}

\begin{figure}
  \centering
  	{\label{fig:us_ranks_curve}\includegraphics[width=0.99\linewidth]{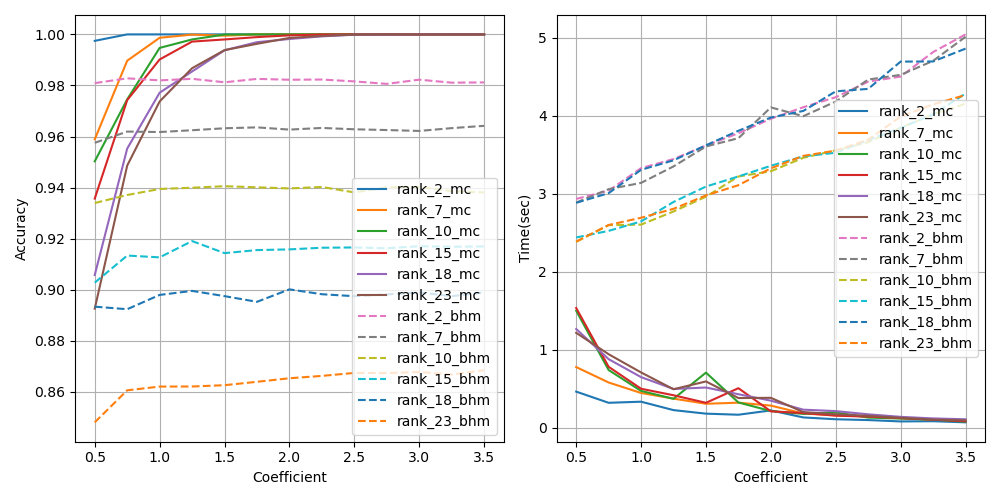}}
  \caption{\small \textit{Left}: TPA performance and \textit{Right}: CT performance per value of $C$ of both methods under NO pattern on different maps with varying values of rank.
  } \vspace{-10pt}
\label{fig:us_ranks_curve}  
\end{figure}

\begin{figure}
  \centering
  	{\label{fig:bus_acc_time_curve}\includegraphics[width=0.99\linewidth]{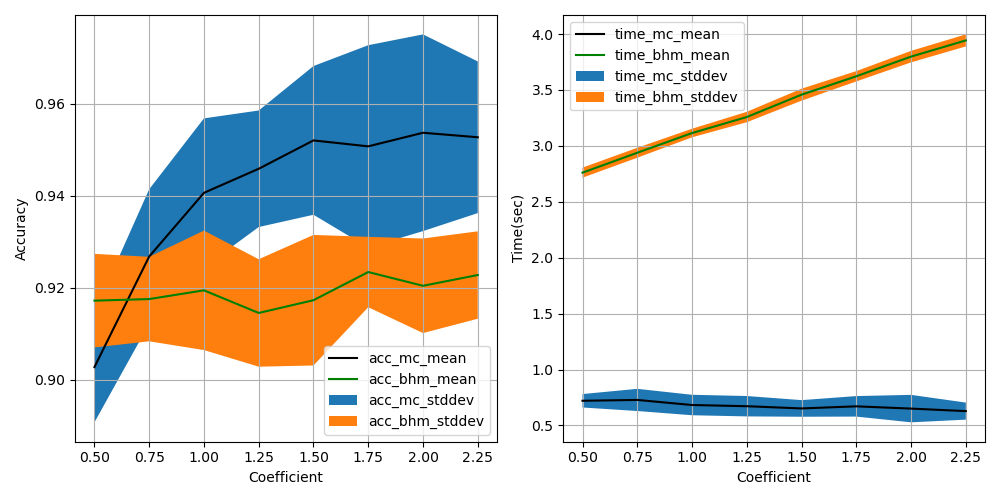}}
  \caption{\small \textit{Left}: Statistical TPA performance and \textit{Right}: CT performance per value of $C$ for LRMC and BHM under PO pattern from 20 different mazes~(with different linear dependencies but the same rank).  
  } \vspace{-10pt}
\label{fig:bus_acc_time_curve}  
\end{figure}

\begin{figure}
  \centering
  	{\label{fig:bus_ranks_curve}\includegraphics[width=0.99\linewidth]{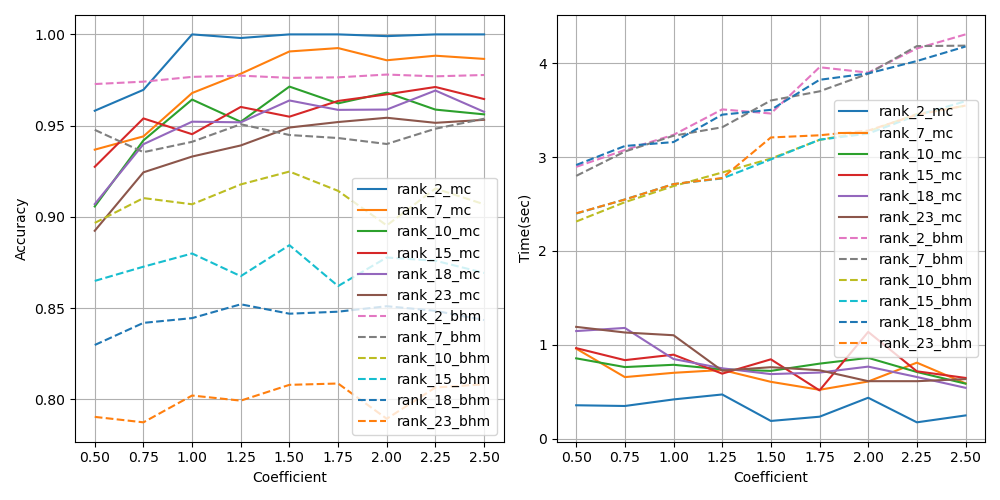}}
  \caption{\small \textit{Left}: TPA performance and \textit{Right}: CT performance per value of $C$ of both methods under PO pattern on different maps with varying values of rank.
  } \vspace{-10pt}
\label{fig:bus_ranks_curve}  
\end{figure}

Note however, from Fig.~\ref{fig:us_acc_time_curve}-\textit{Right} we can see that the LRMC model has a better CT performance than constant time! The reason for this is the LRMC algorithm we use relies on SVD of the matrix. SVD is the most efficient for  dense matrices but not that efficient for sparse ones. In our case,   at first the number of observations can be small so the input matrix is a sparse one. 
The runtime benefit will actually  increase with more observations received, which is a significant feature for real-time robot mapping applications.  

In Fig.~\ref{fig:us_ranks_curve}, we compare the TPA and CT for environments with different ranks. It is clear to see in Fig.~\ref{fig:us_ranks_curve}-\textit{Left} that when the environment becomes more complex~(as the rank value increases), the accuracy of BHM decreases while LRMC maintains the same high-level prediction~(almost $100\%$). From the first row of Fig.~\ref{fig:road_comparison}, we can find BHM generally performs well but many minor discrepancies can be found along the road edges.  When the environment becomes complex, the number of edges increases and thus the minor errors accumulate and this results in the increase of the total map prediction error. The time performances~(Fig.~\ref{fig:us_ranks_curve}-\textit{Right}) of the two methods for different rank-valued maps show the same trend as in Fig.~\ref{fig:us_acc_time_curve}-\textit{Right}.

We then consider the results under PO pattern. Similar to what we have in the NO pattern, the TPA and CT results are shown in Fig.~\ref{fig:bus_acc_time_curve} and Fig.~\ref{fig:bus_ranks_curve}, respectively. The comparison results follow similar trends to the NO pattern. 

\subsection{Coverage Planning with Real-Time Map Prediction}

We investigate if the proposed map prediction can facilitate and improve existing environmental mapping and coverage planning methods. 
Our evaluations are based on three most prevalent strategies. 
The first planning method is the lawnmower (LM) planning: a simple but the most widely used coverage planning method in environmental surveying and monitoring.  Specifically, the traversal patterns and resolutions are pre-determined such that the whole area could be swept incrementally by the robot.
The second planning is the myopic planning (MP) which is typically based on greedy choices (or best-first actions). Here at each action step, a set of points are randomly sampled in all unexplored areas and the closest one to the robot is selected as the local goal. 
The third planning is the non-myopic customized $TSP\_\epsilon$ as described in Section~\ref{sect:plan}, 
where $\epsilon \in \left \{ 0.25,~0.5,~0.75 \right \}$.

\begin{figure*} \vspace{-3pt}
  \centering
  	{\label{fig:road_tsp}\includegraphics[width=0.9\linewidth]{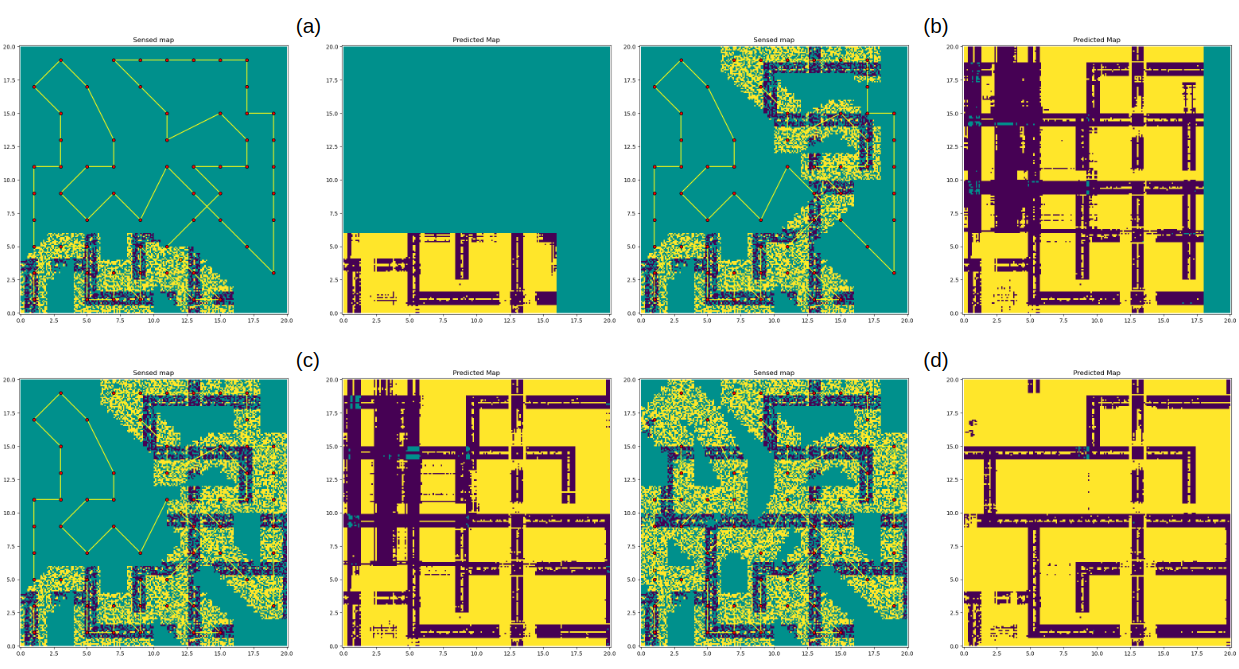}}
  \caption{\small Environmental mapping of an urban road map using TSP\_{0.5}. The path is calculated by using an adaptive \textit{k}-opt TSP planner.
  In each of (a), (b), (c) and (d),  the \textit{Left} figure denotes the observed map while \textit{Right} is the predicted map using LRMC model, respectively. 
  } \vspace{-10pt}
\label{fig:road_tsp}  
\end{figure*}

\begin{figure}[t] 
  \centering
  	{\label{fig:coverage}\includegraphics[width=\linewidth]{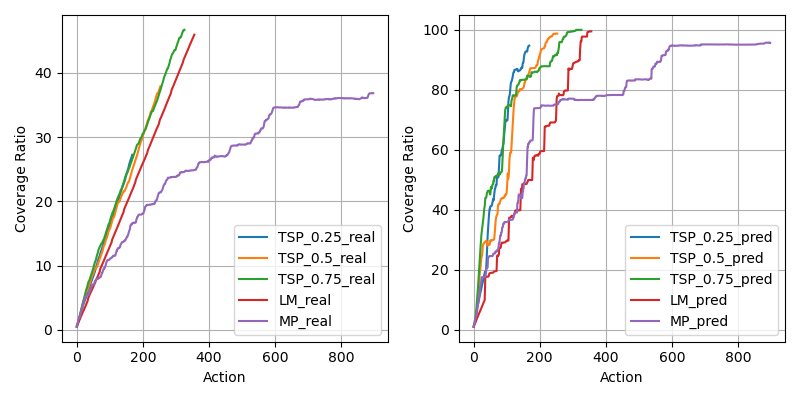}}
  \caption{\small Coverage convergence per action step for different coverage planning methods \textit{Left}: without real-time map prediction and \textit{Right}: with real-time map prediction.
  } \vspace{-10pt}
\label{fig:coverage}  
\end{figure}

The mapping process with non-myopic planning on the urban road network map is demonstrated in Fig.~\ref{fig:road_tsp}, where the real observed (sensed) map and corresponding predicted map from LRMC at four different action steps are presented. The results are consistent with that shown in Fig.~\ref{fig:road_comparison}.

To statistically evaluate the methods, we first consider the coverage planning with only raw observations but no map prediction. (As mentioned earlier, real-time predicted complete maps might not be available due to expensive computation cost in large environments.) The results are shown in Fig.~\ref{fig:coverage}-\textit{Left}, in which none of the listed methods could complete the map coverage within the given number of steps.
This also implies the necessity of real-time map prediction, 
and the results with map prediction are shown in Fig.~\ref{fig:coverage}-\textit{Right}. 
We can see that, 
within the same or less number of steps, nearly all of the planning methods could obtain a complete~(almost $100\%$) mapping coverage with real-time map prediction. 
In addition, the rate for coverage convergence is also remarkably improved for all of the listed coverage planning methods.

According to Fig.~\ref{fig:coverage}-\textit{Right}, among all the coverage planning methods, TSP\_$\epsilon$ methods significantly outperform other methods in terms of coverage convergence, although TSP based method can reach different completeness level as the value of $\epsilon$ varies. 
Empirically a larger value of $\epsilon$ can lead to more observations, a longer path~(and action steps), but also a better level of completeness (i.e., TSP\_{0.75}). A proper choice of $\epsilon$ value can give us a satisfactory trade-off between completeness and required action steps. 
Even with extremely restricted number of action steps~(i.e., less than 200), with a proper $\epsilon$, the TSP based methods can still achieve around $90\%$ coverage. The mapping results at different action steps by TSP\_{0.5} are shown in Fig.~\ref{fig:road_tsp}-(a), (b), (c) and (d), where we can see that our proposed method is able to perform map interpolation and extrapolation simultaneously on the fly during the execution of environmental mapping.

We also list the required number of action steps of all planning methods for different levels of completeness in Table~\ref{tb:coverage_comparison}. It is clear to see that with the proposed real-time LRMC map prediction all coverage planning methods can obtain a substantial improvement of coverage quality, and the required action steps at all listed completeness quantiles are remarkably reduced compared to the results without real-time map prediction.

\begin{table}\vspace{2pt}
\caption{\small Required number of action steps for varying levels of coverage completeness for different methods. The format for the entries indicates \textit{\# without real-time map prediction}~/~\textit{\# with real-time map prediction}, where `-' means the corresponding method fails to finish the respective completeness level.} 
\centering 
{
\begin{tabularx}{0.85\linewidth}{ccccc}
\hline\hline 
   & 20\% & 50\%  & 90\% & 100\% \\[0.5ex]
\hline 
LM  & 153 / 69 & - / 176 & - / 320 &  - / 356~(99\%)\\ [1ex]
MP & 238 / 38 & - / 155 & - / 566 & - / -\\ [1ex]
TSP\_0.25  & 124 / 37 & - / 70 & - / \textbf{151} & - / - \\ [1ex]
TSP\_0.5  & 127 / 19 & - / 101 & - / 195 & - / \textbf{251}~(98\%)\\ [1ex]
TSP\_0.75  & 120  / \textbf{16} & - / \textbf{61} &- / 240 & - / \textbf{309} \\
\hline 
\end{tabularx}
}
\label{tb:coverage_comparison} \vspace{-10pt}
\end{table}

%% file: 5_conclusions.tex
\section{CONCLUSIONS}
We consider the map prediction problem where the measurements are noisy, sparse and partially observed. We first show that many maps possess low-rank and incoherent structures. Then we propose to use LRMC to perform the map prediction. Our extensive experiments validate that the LRMC method can effectively perform map prediction (interpolation and extrapolation), and the LRMC method outperforms  state-of-the-art Bayesian Hilbert Mapping in terms of mapping accuracy and computation time and could update the entire map in real-time. Lastly we show that the proposed real-time map prediction method can be easily combined with popular coverage planning methods and can significantly improve their coverage convergence rates.

%% file: main.bbl
\begin{thebibliography}{10}

\bibitem{senanayake2017bayesian}
Ransalu Senanayake and Fabio Ramos.
\newblock Bayesian hilbert maps for dynamic continuous occupancy mapping.
\newblock In {\em Conference on Robot Learning}, pages 458--471, 2017.

\bibitem{ramos2016hilbert}
Fabio Ramos and Lionel Ott.
\newblock Hilbert maps: scalable continuous occupancy mapping with stochastic
  gradient descent.
\newblock {\em The International Journal of Robotics Research},
  35(14):1717--1730, 2016.

\bibitem{elfes1989using}
Alberto Elfes.
\newblock Using occupancy grids for mobile robot perception and navigation.
\newblock {\em Computer}, 22(6):46--57, 1989.

\bibitem{o2012gaussian}
Simon~T O’Callaghan and Fabio~T Ramos.
\newblock Gaussian process occupancy maps.
\newblock {\em The International Journal of Robotics Research}, 31(1):42--62,
  2012.

\bibitem{wang2016fast}
Jinkun Wang and Brendan Englot.
\newblock Fast, accurate gaussian process occupancy maps via test-data octrees
  and nested bayesian fusion.
\newblock In {\em 2016 IEEE International Conference on Robotics and Automation
  (ICRA)}, pages 1003--1010. IEEE, 2016.

\bibitem{jadidi2014exploration}
Maani~Ghaffari Jadidi, Jaime~Valls Mir{\'o}, Rafael Valencia, and Juan
  Andrade-Cetto.
\newblock Exploration on continuous gaussian process frontier maps.
\newblock In {\em 2014 IEEE International Conference on Robotics and Automation
  (ICRA)}, pages 6077--6082. IEEE, 2014.

\bibitem{jadidi2015mutual}
Maani~Ghaffari Jadidi, Jaime~Valls Miro, and Gamini Dissanayake.
\newblock Mutual information-based exploration on continuous occupancy maps.
\newblock In {\em 2015 IEEE/RSJ International Conference on Intelligent Robots
  and Systems (IROS)}, pages 6086--6092. IEEE, 2015.

\bibitem{caley2019deep}
Jeffrey~A Caley, Nicholas~RJ Lawrance, and Geoffrey~A Hollinger.
\newblock Deep learning of structured environments for robot search.
\newblock {\em Autonomous Robots}, 43(7):1695--1714, 2019.

\bibitem{pronobis2017learning}
Andrzej Pronobis and Rajesh~PN Rao.
\newblock Learning deep generative spatial models for mobile robots.
\newblock In {\em 2017 IEEE/RSJ International Conference on Intelligent Robots
  and Systems (IROS)}, pages 755--762. IEEE, 2017.

\bibitem{katyal2018occupancy}
Kapil Katyal, Katie Popek, Chris Paxton, Joseph Moore, Kevin Wolfe, Philippe
  Burlina, and Gregory~D Hager.
\newblock Occupancy map prediction using generative and fully convolutional
  networks for vehicle navigation.
\newblock {\em arXiv preprint arXiv:1803.02007}, 2018.

\bibitem{shrestha2019learned}
Rakesh Shrestha, Fei-Peng Tian, Wei Feng, Ping Tan, and Richard Vaughan.
\newblock Learned map prediction for enhanced mobile robot exploration.
\newblock In {\em 2019 International Conference on Robotics and Automation
  (ICRA)}, pages 1197--1204. IEEE, 2019.

\bibitem{saroya2020online}
Manish Saroya, Graeme Best, and Geoffrey~A Hollinger.
\newblock Online exploration of tunnel networks leveraging topological
  cnn-based world predictions.
\newblock In {\em Proc. of IEEE/RSJ IROS}, 2020.

\bibitem{katyal2019uncertainty}
Kapil Katyal, Katie Popek, Chris Paxton, Phil Burlina, and Gregory~D Hager.
\newblock Uncertainty-aware occupancy map prediction using generative networks
  for robot navigation.
\newblock In {\em 2019 International Conference on Robotics and Automation
  (ICRA)}, pages 5453--5459. IEEE, 2019.

\bibitem{candes2009exact}
Emmanuel~J Cand{\`e}s and Benjamin Recht.
\newblock Exact matrix completion via convex optimization.
\newblock {\em Foundations of Computational mathematics}, 9(6):717, 2009.

\bibitem{candes2010power}
Emmanuel~J Cand{\`e}s and Terence Tao.
\newblock The power of convex relaxation: Near-optimal matrix completion.
\newblock {\em IEEE Transactions on Information Theory}, 56(5):2053--2080,
  2010.

\bibitem{chen2015incoherence}
Yudong Chen.
\newblock Incoherence-optimal matrix completion.
\newblock {\em IEEE Transactions on Information Theory}, 61(5):2909--2923,
  2015.

\bibitem{gross2011recovering}
David Gross.
\newblock Recovering low-rank matrices from few coefficients in any basis.
\newblock {\em IEEE Transactions on Information Theory}, 57(3):1548--1566,
  2011.

\bibitem{recht2011simpler}
Benjamin Recht.
\newblock A simpler approach to matrix completion.
\newblock {\em Journal of Machine Learning Research}, 12(12), 2011.

\bibitem{liu2010interior}
Zhang Liu and Lieven Vandenberghe.
\newblock Interior-point method for nuclear norm approximation with application
  to system identification.
\newblock {\em SIAM Journal on Matrix Analysis and Applications},
  31(3):1235--1256, 2010.

\bibitem{cai2010singular}
Jian-Feng Cai, Emmanuel~J Cand{\`e}s, and Zuowei Shen.
\newblock A singular value thresholding algorithm for matrix completion.
\newblock {\em SIAM Journal on optimization}, 20(4):1956--1982, 2010.

\bibitem{ji2009accelerated}
Shuiwang Ji and Jieping Ye.
\newblock An accelerated gradient method for trace norm minimization.
\newblock In {\em Proceedings of the 26th annual international conference on
  machine learning}, pages 457--464, 2009.

\bibitem{mazumder2010spectral}
Rahul Mazumder, Trevor Hastie, and Robert Tibshirani.
\newblock Spectral regularization algorithms for learning large incomplete
  matrices.
\newblock {\em The Journal of Machine Learning Research}, 11:2287--2322, 2010.

\bibitem{liu2017new}
Guangcan Liu, Qingshan Liu, and Xiaotong Yuan.
\newblock A new theory for matrix completion.
\newblock In {\em Advances in Neural Information Processing Systems}, pages
  785--794, 2017.

\bibitem{bhojanapalli2014universal}
Srinadh Bhojanapalli and Prateek Jain.
\newblock Universal matrix completion.
\newblock {\em arXiv preprint arXiv:1402.2324}, 2014.

\bibitem{chen2014coherent}
Yudong Chen, Srinadh Bhojanapalli, Sujay Sanghavi, and Rachel Ward.
\newblock Coherent matrix completion.
\newblock In {\em International Conference on Machine Learning}, pages
  674--682, 2014.

\bibitem{eriksson2012high}
Brian Eriksson, Laura Balzano, and Robert Nowak.
\newblock High-rank matrix completion.
\newblock In {\em Artificial Intelligence and Statistics}, pages 373--381,
  2012.

\bibitem{chen2015completing}
Yudong Chen, Srinadh Bhojanapalli, Sujay Sanghavi, and Rachel Ward.
\newblock Completing any low-rank matrix, provably.
\newblock {\em The Journal of Machine Learning Research}, 16(1):2999--3034,
  2015.

\bibitem{eftekhari2018mc2}
Armin Eftekhari, Michael~B Wakin, and Rachel~A Ward.
\newblock Mc2: A two-phase algorithm for leveraged matrix completion.
\newblock {\em Information and Inference: A Journal of the IMA}, 7(3):581--604,
  2018.

\bibitem{nguyen2019low}
Luong~Trung Nguyen, Junhan Kim, and Byonghyo Shim.
\newblock Low-rank matrix completion: A contemporary survey.
\newblock {\em IEEE Access}, 7:94215--94237, 2019.

\bibitem{recht2009simpler}
Benjamin Recht.
\newblock A simpler approach to matrix completion.
\newblock 2009.

\bibitem{ongie2018tensor}
Greg Ongie, Daniel Pimentel-Alarc{\'o}n, Laura Balzano, Rebecca Willett, and
  Robert~D Nowak.
\newblock Tensor methods for nonlinear matrix completion.
\newblock {\em arXiv preprint arXiv:1804.10266}, 2018.

\bibitem{laurent2014positive}
Monique Laurent and Antonios Varvitsiotis.
\newblock Positive semidefinite matrix completion, universal rigidity and the
  strong arnold property.
\newblock {\em Linear Algebra and its Applications}, 452:292--317, 2014.

\bibitem{li2019coverage}
Teng Li, Chaoqun Wang, Max Q-H Meng, and Clarence~W de~Silva.
\newblock Coverage sampling planner for uav-enabled environmental exploration
  and field mapping.
\newblock {\em arXiv preprint arXiv:1907.05910}, 2019.

\bibitem{song2018varepsilon}
Junnan Song and Shalabh Gupta.
\newblock An online coverage path planning algorithm.
\newblock {\em IEEE Transactions on Robotics}, 34(2):526--533, 2018.

\bibitem{khan2014greedy}
Fahad~Ahmad Khan, Saad~Ahmad Khan, Damla Turgut, and Ladislau B{\"o}l{\"o}ni.
\newblock Greedy path planning for maximizing value of information in
  underwater sensor networks.
\newblock In {\em 39th Annual IEEE Conference on Local Computer Networks
  Workshops}, pages 610--615. IEEE, 2014.

\bibitem{kantaros2019asymptotically}
Yiannis Kantaros, Brent Schlotfeldt, Nikolay Atanasov, and George~J Pappas.
\newblock Asymptotically optimal planning for non-myopic multi-robot
  information gathering.
\newblock In {\em Robotics: Science and Systems}, 2019.

\bibitem{ma2016information}
Kai-Chieh Ma, Lantao Liu, and Gaurav~S Sukhatme.
\newblock An information-driven and disturbance-aware planning method for
  long-term ocean monitoring.
\newblock In {\em 2016 IEEE/RSJ International Conference on Intelligent Robots
  and Systems (IROS)}, pages 2102--2108. IEEE, 2016.

\bibitem{chen2019pareto}
Weizhe Chen and Lantao Liu.
\newblock Pareto monte carlo tree search for multi-objective informative
  planning.
\newblock In {\em Robotics: Science and Systems}, 2019.

\bibitem{ma2016adaptive}
Zhibei Ma, Lantao Liu, and Gaurav~S Sukhatme.
\newblock An adaptive k-opt method for solving traveling salesman problem.
\newblock In {\em 2016 IEEE 55th Conference on Decision and Control (Cdc)},
  pages 6537--6543. IEEE, 2016.

\bibitem{daedalus}
Walter~D. Pullen.
\newblock Daedalus program.
\newblock Available at \url{http://www.astrolog.org/labyrnth/daedalus.htm}.

\end{thebibliography}
